\documentclass[10pt,twocolumn,letterpaper]{article}

\usepackage{cvpr}
\usepackage[colorlinks,linkcolor=red]{hyperref}
\usepackage{fancyhdr}
\usepackage{times}
\usepackage{epsfig}
\usepackage{graphicx}
\usepackage{amsmath}
\usepackage{amssymb}
\usepackage{subcaption}
\usepackage{booktabs}
\usepackage{multirow}
\usepackage{diagbox}
\usepackage{color}
\usepackage[title]{appendix}

\cvprfinalcopy 

\def\cvprPaperID{****} 

\begin{document}

\title{Discriminative Multi-modality Speech Recognition}

\author{Bo Xu, Cheng Lu, Yandong Guo and Jacob Wang\\
Xpeng motors\\
{\tt\small xiaoboboer@gmail.com}
}

\maketitle
\begin{abstract}
Vision is often used as a complementary modality for audio speech recognition (ASR), especially in the noisy environment where performance of solo audio modality significantly deteriorates. After combining visual modality, ASR is upgraded to the multi-modality speech recognition (MSR). In this paper, we propose a two-stage speech recognition model. In the first stage, the target voice is separated from background noises with help from the corresponding visual information of lip movements, making the model `listen' clearly. At the second stage, the audio modality combines visual modality again to better understand the speech by a MSR sub-network, further improving the recognition rate. There are some other key contributions: we introduce a pseudo-3D residual convolution (P3D)-based visual front-end to extract more discriminative features; we upgrade the temporal convolution block from 1D ResNet with the temporal convolutional network (TCN), which is more suitable for the temporal tasks; the MSR sub-network is built on the top of Element-wise-Attention Gated Recurrent Unit (EleAtt-GRU), which is more effective than Transformer in long sequences. We conducted extensive experiments on the LRS3-TED and the LRW datasets. Our two-stage model (audio enhanced multi-modality speech recognition, AE-MSR) consistently achieves the state-of-the-art performance by a significant margin, which demonstrates the necessity and effectiveness of AE-MSR.
   
\end{abstract}

\section{Introduction}

In the book \emph{The Listening Eye: A Simple Introduction to the Art of Lip-reading}~\cite{clegg1953listening}, Clegg mentions that ``When you are deaf you live inside a well-corked glass bottle. You see the entrancing outside world, but it does not reach you. After learning to lip read, you are still inside the bottle, but the cork has come out and the outside world slowly but surely comes in to you."
Lip reading is an approach for people with hearing impairments to communicate with the world, so that they can interpret what other say by looking at the movements of lips~\cite{assael2016lipnet,chung2016out,fisher1968confusions,mcgurk1976hearing,woodward1960phoneme}. Lip reading is a difficult skill for human to grasp and requires intensive training~\cite{easton1982perceptual,stafylakis2017combining}. Lip reading is also an inexact art, because different characters may exhibit the similar lip movements (\emph{e.g.} `b' and `p')~\cite{Triantafyllos-avsr2018}. Consequently, several machine lip reading models are proposed to discriminate such subtle difference~\cite{cooke2006audio,kumar2007profile,petajan1985automatic}. However they still suffer difficulties on extracting spatio-temporal features from the video.

\begin{figure}[t]
\centering
\includegraphics[width=1.0\linewidth]{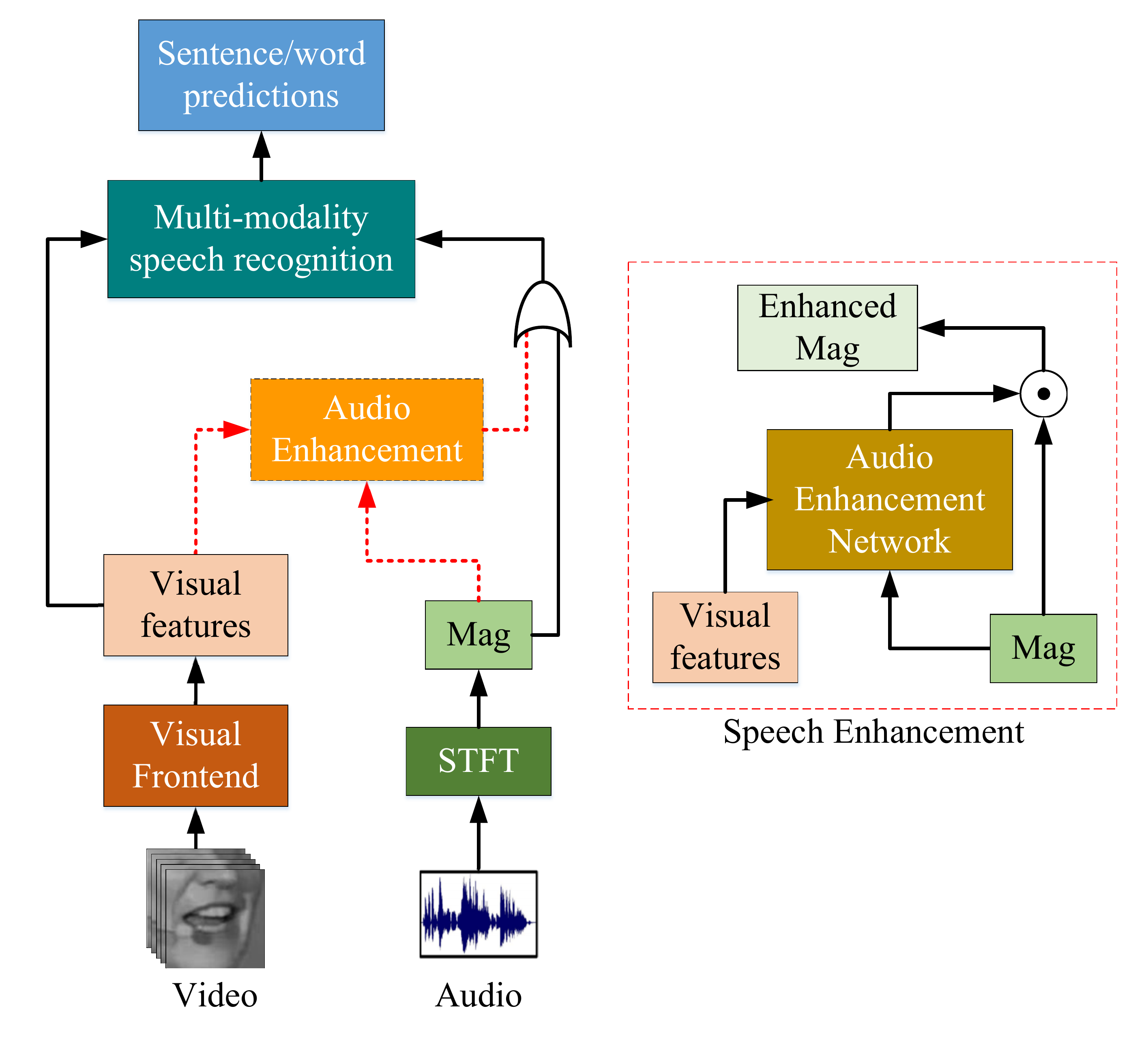}
\caption{Overview of the audio enhanced multi-modality speech recognition network (AE-MSR). {\bf Mag:} magnitude. Different from other MSR methods~\cite{Triantafyllos-avsr2018,chung2017lip,yang2019lrw,stafylakis2017combining,petridis2018end} with only single visual awareness, we firstly filter the voices of speakers and background noises with help from visual awareness. Then we combine visual awareness again for MSR to benefit speech recognition.}
\label{fig:long}
\label{fig:onecol}
\end{figure}
Automatic lip reading becomes achievable due to rapid development of deep neural network in computer vision~\cite{krizhevsky2012imagenet,simonyan2014very,szegedy2015going}, and with help from large scale training datasets~\cite{chung2017lip,chung2016lip,cooke2006audio,czyzewski2017audio,russakovsky2015imagenet,yang2019lrw}. In addition to serving as a powerful solution to hearing impairment, lip reading can also contribute to audio speech recognition (ASR) in adversary environments, such as in high noise level where human speaking is inaudible. Multi-modality (video and audio) is more effective than single modality (video or audio) in terms of both robustness and accuracy. Multi-modality (audio-visual) speech recognition (MSR) is one of the main extended applications of multi-modality. But similar to ASR, there is a significant deterioration in performance for MSR in noisy environment as well~\cite{Triantafyllos-avsr2018}. Compared to audio modality operating in a clean voice environment, the one in noisy environment shows less gain because of upgrading from ASR to MSR.~\cite{Triantafyllos-avsr2018} demonstrates that the noisy level of audio modality directly impacts the performance gain of MSR compared to single modality.

The goal of this paper is to introduce a two-stage speech recognition method with double visual-modality awareness. In the first stage, we reconstruct the audio signal which only contains the target speaker's voice with the guiding visual information (lip movements). In the second stage, the enhanced audio modality is combined with the visual modality again to yield better speech recognition. Compared to typical MSR methods with single time of visual modality awareness, our method is more advantageous in terms of robustness and accuracy.

We propose a deep neural network model named audio-enhanced multi-modality speech recognition (AE-MSR) with double visual awareness to implement the method. The AE-MSR model consists of two sub-networks, the audio enhancement (AE) and MSR. Before being fed into the MSR sub-network, audio modality is enhanced with help from the first visual awareness in the AE sub-network. After enhancement, audio stream and revisited visual stream are then fed into the MSR sub-network to make speech predictions.The techniques we incorporated into AE-MSR include pseudo 3D residual convolution (P3D), temporal convolutional network (TCN), and element-wise attention gated recurrent unit (EleAtt-GRU). Ablation study shown in the paper demonstrates the effectiveness of each of the above sub-modules and the combination of them. The MSR sub-network is also built on top of EleAtt-GRU. 

The intuition of our AE-MSR is as follows. Typically, a deep learning-based MSR uses symmetric encoders for both audio and video. Though visual encoder is trained in an e2e fashion, we conduct experiments to show this is not the optimal way to leverage the visual information. The reason might be that the intrinsic architecture of the typical MSR implicitly suggests equal importance of audio and video. However we tell from various experiments that audio is still much more reliable to recognize speech, even in a noisy environment. Therefore, we re-design the architecture to embed this bias between video and audio as a prior. 
\\

Overall, the contributions of this paper are:
\begin{itemize}
    \item We propose a two-stage double visual awareness MSR model, where the first visual awareness is applied to remove the audio noise.  
    \item We introduce the P3D as visual front-end to extract more discriminative visual features and EleAtt-GRU to adaptively encode the spatio-temporal information in AE and MSR sub-networks, benefiting performance of both networks.
    \item We upgrade the temporal convolution block of 1D ResNet to a TCN one in AE sub-network for establishing temporal connections.
    \item Extensive experiments demonstrate that AE-MSR surpasses state-of-the-art MSR model~\cite{Triantafyllos-avsr2018} both in audio clean and noisy environments on the Lip Reading Sentences 3 (LRS3-TED) dataset~\cite{afouras2018lrs3}. The word classification model we build based on P3D also outperforms the word-level state-of-the-art~\cite{stafylakis2017combining} on the Lip Reading in the Wild (LRW) dataset~\cite{chung2016lip}. 
\end{itemize}

\section{Related works}

In this section, we introduce some related works about audio enhancement (AE) driven by visual information and multi-modality speech recognition (MSR).

\begin{figure*}[t]
\centering
    \begin{subfigure}[t]{6.2in}
        \centering
        \includegraphics[width=0.85\linewidth]{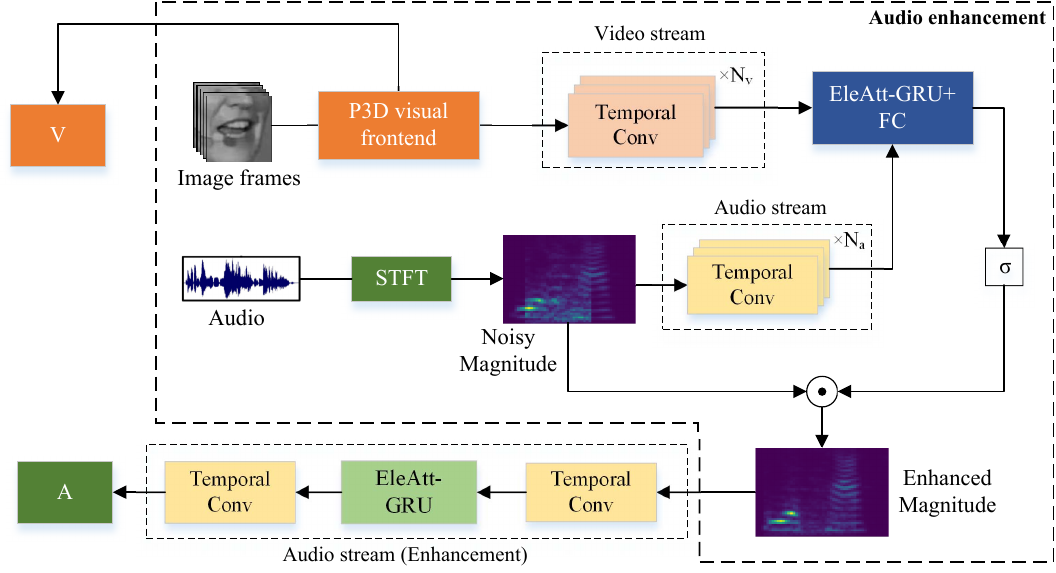}
        \caption{Audio enhancement (AE)
        sub-network}
        \label{fig:architecture_ae}
    \end{subfigure}
    
    \begin{subfigure}[h]{6.2in}
        \centering
        \includegraphics[width=1.05\linewidth]{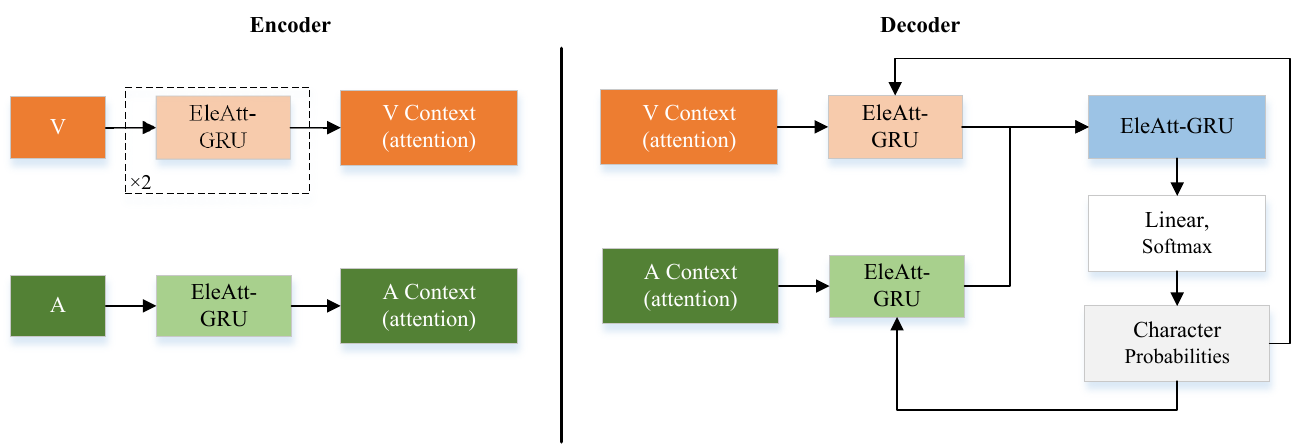}
        \caption{Multi-modality speech recognition (MSR) sub-network}
        \label{fig:architecture_MSR}
    \end{subfigure}
\caption{Architecture of the multi-modality speech recognition network with double visual awareness ({\bf AE-MSR}). The AE-MSR network consists of two sub-networks: {\bf a)} the audio enhancement (AE) network. The network receives image frames and audio signals as inputs, outputting the enhanced magnitude spectrograms that the noisy spectrograms are filtered. {\bf V:} visual features; {\bf A:} enhanced audio magnitude. {\bf b)} the multi-modality speech recognition (MSR) network.}
\label{fig:architecture}
\vspace{6pt}
\end{figure*}

\subsection{Audio enhancement}

A few researchers have demonstrated that the target audio signal can be separated from other speakers' voices and background noises, \eg Gabbay \etal~\cite{gabbay2018seeing} introduce a trained silent-video-to-speech model previously proposed by~\cite{ephrat2017improved} to generate speech predictions as a mask on the noisy audio signal which is then discarded in the pipeline of audio enhancement. Gabbay \etal~\cite{gabbay2017visual} also use the convolution neural networks (CNNs) to encode multi-modality features. The embedding vectors of audio and vision are concatenated before audio decoder and fed into transposed convolution of audio decoder to produce enhanced mel-scale spectrograms. Hou \etal~\cite{hou2018audio} build a visual driven AE network on the top of CNNs and fully connected (FC) layers to generate enhanced speech and reconstructed lip image frames. Afouras ~\etal~\cite{afouras2018conversation} use 1D ResNet as temporal convolution unit to process audio and visual features individually. Then the multi-modality features are concatenated and encoded into a mask by another 1D-ResNet-based encoder to remove noisy components in the audio signal. In their latest article, they propose a new approach that replaces the multi-modality feature encoder with Bi-LSTM~\cite{afouras2019my}.

\subsection{Multi-modality speech recognition}

Vision is often used as a complementary modality for audio speech recognition (ASR), especially in noisy environments. After combining visual modality, ASR is upgraded to the multi-modality speech recognition (MSR). Reciprocally, MSR is also an upgrade to the lip reading and benefits people with hearing impairments to recognize speech by generating meaningful text.

In the field of deep learning, research on lip reading has longer history than MSR~\cite{zhou2014review}. Assael \etal~\cite{assael2016lipnet} propose LipNet, an end-to-end model on top of spatio-temporal convolutions, LSTM~\cite{hochreiter1997long} and connectionist temporal classification (CTC) loss on variable-length sequence of video frames. Stafylakis \etal~\cite{stafylakis2017combining} introduce the state-of-the-art word-level classification lip reading network on LRW dataset~\cite{chung2016lip}. The network consists of spatio-temporal convolution, residual network and Bi-LSTM. 

On the basis of lip reading, MSR is developed~\cite{chung2017lip,Triantafyllos-avsr2018}. Various MSR methods often use encoder-to-decoder (enc2dec) mechanism which is inspired by machine translation~\cite{bahdanau2014neural,chan2016listen,graves2006connectionist,graves2014towards,sutskever2014sequence,vaswani2017attention}. Chung \etal~\cite{chung2017lip} use a dual sequence-to-sequence model with enc2dec mechanism. Visual features and audio features are encoded separately by LSTM units. Then multi-modality features are combined and decoded into characters. Afouras \etal~\cite{Triantafyllos-avsr2018} introduce a sequence-to-sequence model of encoder-to-decoder mechanism. The encoder and decoder of the model are built based on the transformer~\cite{vaswani2017attention} attention architecture. In encoder stage, each modality feature is encoded with self-attention individually. After multi-head attention in decoder stage, the context vectors produced by multiple modalities are concatenated and fed to the feed forward layers to produce probable characters. However, their state-of-the-art method suffer in noisy scenarios. In noisy environments, the performance dramatically decreases, this is the main reason why we propose the method of AE-MSR. In this paper, we qualitatively evaluate performance of the AE-MSR model for speech recognition in the noisy environments.  
\section{Architectures}
In this section, we describe the double visual awareness multi-modality speech recognition (AE-MSR) network. It first learns to filter magnitude spectrogram from the voices of other speakers or background noises with help from the information of visual modality (\emph{Watch once to listen clearly}). The subsequent MSR then revisits the visual modality and combines it with filtered audio magnitude spectrogram  (\emph{Watch again to understand precisely}). The model architecture is shown in detail in Figure~\ref{fig:architecture}.
\begin{figure}[t]
    \begin{subfigure}[t]{1.4in}
        \centering
        \includegraphics[width=1.5\linewidth]{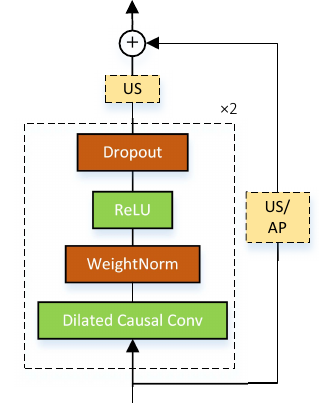}
        \caption{TCN ResNet block}
        \label{fig:TCN_ResNet_block}
    \end{subfigure}
    \qquad\qquad
    \begin{subfigure}[t]{1.4in}
        \centering
        \includegraphics[width=1.1\linewidth]{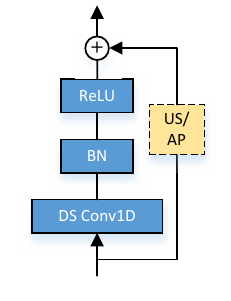}
        \caption{1D ResNet block}
        \label{fig:1D_ResNet_block}
    \end{subfigure}
\caption{Temporal convolution blocks.  {\bf a) TCN ResNet block}. {\bf US:} Up-sample; {\bf AP:} Average Pooling~\cite{he2016deep}. {\bf b) The 1D ResNet block}. {\bf DS:} Depthwise separable~\cite{chollet2017xception}; {\bf BN:} Batch Normalization. The non-upsample convolution layers are all depthwise separable.}
\label{fig:temporal_conv}
\end{figure}\\
\subsection{Watch once to listen clearly}
{\bf Audio features.} We use Short Time Fourier Transform (STFT) to extract magnitude spectrogram from the waveform signal at a sample rate of 16kHz. To align with the video frame rate at 25fps, we set the STFT window length to 40ms and hop length to 10ms, corresponding to 75\% overlap. We multiply the resulting magnitude by a mel-spaced filter to compute the audio feature of mel-scale magnitude with mel-frequency bins of 80 between 0 to 8 kHz. 

{\bf Visual features.} We produce image frames by cropping original video frames to $112 \times 112$ pixel patches and choose mouth patch as region of interest (ROI). To extract video features, we build a 3D CNN (C3D)~\cite{tran2015learning} -P3D~\cite{qiu2017learning} network to produce a more powerful visual spatio-temporal representation instead of using C3D plus 2D ResNet~\cite{he2016deep} which is mentioned in many other lip-reading papers~\cite{Triantafyllos-avsr2018,afouras2018conversation,afouras2018deep,afouras2019my,chung2017lip,stafylakis2017combining}.

C3D is a beneficial method to capture spatio-temporal features of videos and widely adopted~\cite{stafylakis2017combining,Triantafyllos-avsr2018,afouras2018conversation,petridis2018end,afouras2019my}. Multi-layer C3D can achieve even better performances in temporal tasks than a single layer one, however they are both computationally expensive and memory demanding. We use P3D to replace part of the C3D layers to alleviate this situation. The three block versions of P3D are shown in Figure~\ref{fig:P3D}, P3D ResNet is implemented by separating $N\times N\times N$ convolutions into $1\times3\times3$ convolution filters on spatial domain and $3\times1\times1$ convolution filters on temporal domain to extract spatial-temporal features. P3D ResNet achieves superior performances over 2D ResNet in different temporal tasks~\cite{qiu2017learning}. We implement a 50-layer P3D network by cyclically mixing the three blocks in the order of P3D-A, P3D-B, P3D-C.

The visual front-end is built on a 3D convolution layer with 64 filters of kernel size $5\times7\times7$, followed by batch normalization (BN), ReLU activation and max-pooling layers. And then the max-pooling is followed by a 50-layer P3D ResNet that gradually decreases the spatial dimensions with depth while maintaining the temporal dimension. For an input of $T\times H\times W$ frames, the output of the sub-network is a $T\times512$ tensor (in the final stage, the feature is average-pooled in spatial dimension and processed as a 512-dimensional vector representing each video frame). The visual feature and corresponding magnitude spectrogram are then fed into audio enhancement sub-network.

{\bf Audio enhancement with the first visual awareness.} Noise-free audio signal achieves satisfactory performance on audio speech recognition (ASR) and multi-modality speech recognition (MSR). However there is a significant deterioration in recognition performance in noisy environments~\cite{Triantafyllos-avsr2018,afouras2018conversation}. Architecture of the audio enhancement sub-network is illustrated in Figure~\ref{fig:architecture_ae}, where the visual features are fed into a temporal convolution network (video stream). The video stream consists of $N_v$ temporal convolution blocks, outputting video feature vectors. We introduce two versions of temporal convolution blocks, one is the temporal convolutional network (TCN) proposed by~\cite{bai2018empirical} and the other is 1D ResNet block proposed by~\cite{afouras2019my}.

Architectures of temporal convolution blocks are shown in Figure~\ref{fig:temporal_conv}, the residual block of TCN consists of two dilated causal convolution layers, each layer followed by a weight normalization (WN)~\cite{salimans2016weight} layer and a rectified linear unit (ReLU)~\cite{nair2010rectified} layer. A spatial dropout~\cite{srivastava2014dropout} layer is added after ReLU layer for regularization~\cite{bai2018empirical}. Identity skip connection are added after the second dilated causal convolution layer. By combining causal convolution and dilated convolution, TCN guarantees no leakage from the future to the past and effectively expands the receptive field to maintain a longer memory size~\cite{bai2018empirical}. The 1D ResNet block is based on 1D temporal convolution layer, followed by a batch normalization (BN) layer. Residual connection is added after ReLU activation layer. 

Two of the intermediate temporal convolution blocks containing transposed convolution layers up-sample the video features by 4 to match the temporal dimension of the audio feature vectors (4$T$). Similarly, the noisy magnitude spectrograms are proposed by a residual network (audio stream) which consists of $N_a$ temporal convolution blocks, outputting audio feature vectors. Then the audio feature vectors and the video feature vectors are fused in a fusion layer by simply concatenating over the channel dimension. The fused multi-modality vector is then fed into a one-layer EleAtt-GRU encoder followed by 2 fully connected layers with a Sigmoid as activation to produce a target enhancement mask (values range from 0 to 1). EleAtt-GRU is demonstrated more effective than other RNN variants in spatio-temporal tasks and its detail is introduced in section~\ref{Multi-modality speech recognition}. The enhanced magnitude is produced by multiplying the original magnitude spectrogram with the target enhancement mask element-wise. Architecture details of the audio enhancement sub-network are given in Table~\ref{enhancement_details}.

\subsection{Watch again to understand precisely} \label{Multi-modality speech recognition}
{\bf Multi-modality speech recognition with the second visual awareness.} Visual information can help enhance audio modality by separating target audio signal from noisy background. After audio enhancement by visual awareness, the multi-modality speech recognition (MSR) is implemented by combining enhanced audio and the revisited visual representation to benefit the performance of speech recognition further.

We use encoder-to-decoder (enc2dec) mechanism in the MSR sub-network. Instead of using transformer~\cite{vaswani2017attention}, which demonstrates decent performance on lip reading~\cite{afouras2018deep} and MSR~\cite{Triantafyllos-avsr2018}, our network is basically built on a RNN variant model named gated recurrent unit with element-wise-attention (EleAtt-GRU)~\cite{zhang2019eleatt}. Although transformer is a powerful model emerging in machine translation~\cite{vaswani2017attention} and lip reading~\cite{Triantafyllos-avsr2018,afouras2018deep}, it builds character relationships within limited length, less effective with long sequences than RNN. EleAtt-GRU can alleviate this situation, because it is equipped with an element-wise-attention gate (EleAttG) that empowers an RNN neuron to have the attentive capacity. EleAttG is designed to modulate the input adaptively by assigning different importance levels, i.e., attention, to each element or dimension of the input. Illustration of EleAttG for a GRU block is shown in Figure~\ref{fig:EleAtt-GRU}. In a GRU block/layer, all neurons share the same EleAttG, which reduces the cost of computation and number of parameters.    

Architecture of the AE-MSR network is shown in Figure~\ref{fig:architecture}, a sequence-to-sequence MSR network is built based on EleAtt-GRU. The encoder is a two-layer EleAtt-GRU for both modalities. The enhanced audio magnitude is fed into an encoder layer between two 1D-ResNet blocks with stride 2 that down-sample the temporal dimension by 4 to match the temporal dimension of video features ($T$). The 1D-ResNet layer are followed by another encoder layer, outputting the audio modality encoder context. The video features extracted by C3D-P3D network are fed into the video encoder to output video encoder context. In the decoder stage, video context and audio context are decoded separately by independent decoder layer. Generated context vectors of both modalities are concatenated over the channel dimensions and propagated to another decoder layer to produce character probabilities. The number of unit of EleAtt-GRU in both encoder and decoder is 128. The decoder outputs character probabilities which are directly matched to the ground truth labels and trained with a cross-entropy loss and the whole output sequence is trained with sequence-to-sequence (seq2seq) loss~\cite{sutskever2014sequence}.
\section{Training}

\subsection{Datasets}

The proposed network is trained and evaluated on LRW~\cite{chung2016lip} and LRS3-TED~\cite{afouras2018lrs3} datasets. LRW is a very large-scale lip reading dataset in the wild from British television broadcasts, including news and talk shows. LRW consists of up to 1000 utterances of 500 different words, spoken by more than 1000 speakers. We use LRW dataset to pre-train the P3D spatio-temporal front-end based on a word-level classification network of lip reading.  

LRS3-TED is the largest available dataset in the field of lip reading (visual speech recognition). It consists of face tracks from over 400 hours of TED and TEDx videos, and organized into three sets: pre-train, train-val and test. We train the audio enhancement (AE) sub-network and the multi-modality speech recognition (MSR) sub-network on the LRS3-TED dataset.
\subsection{Evaluation metric}
For the word-level lip reading experiment, the train, validation and test sets are provided with the LRW dataset. We report word accuracy for classification in 500 word classes of LRW. For sentence-level recognition experiments, we report the Word Error Rate (WER). WER is defined as $WER = (S+D+I)/N$, where $S$ is the number of substitution, $D$ is the number of deletions, $I$ is the number of insertions to get from the reference to the hypothesis, and $N$ is the number of words in the reference~\cite{chung2017lip}. 

\subsection{Training strategy}\label{training_strategy}
{\bf Visual front-end.} The visual front-end of C3D-P3D is pre-trained on a word-level classification network of lip reading with LRW dataset for 500 word classes and we adopt a two-step training strategy. In the first step, image frames are fed into a 3D convolution, which is followed by a 50-layer P3D, and the back-end is based on one dense layer. In the second step, to improve the effectiveness of the model we replace the dense layer with two layers of Bi-LSTM, followed by a linear and a SoftMax layer. We use cross entropy loss to train the word classification tasks. With the visual front-end frozen, we extract and save video features, as well as magnitude spectrograms for both original audio and the mix-noise one.

{\bf Noisy samples.} In order to train our model so that it can be resistant to background noise or speakers, we follow the noise mix method of~\cite{Triantafyllos-avsr2018}, the babble noise with SNR from -10 dB to 10 dB is added to audio stream with probability p$_n$ = 0.25 and the babble noise samples are synthesized by mixing the signals of 30 different audio samples from LRS3-TED dataset. 

{\bf AE and MSR sub-networks.} The AE sub-network is firstly trained on multi-modality of mixed noises with temporal convolution block of TCN and 1D ResNet separately. The AE sub-network is trained by minimizing the \emph{L}$_1$ loss between the predicted magnitude spectrogram and the ground truth. Simultaneously, the multi-modality speech recognition (MSR) sub-network is trained with video features and clean magnitude spectrogram as inputs. The MSR sub-network is also trained when only single modality (audio or visual) is available. For MSR sub-network, we use a sequence-to-sequence (seq2seq) loss~\cite{cho2014learning,sutskever2014sequence}.

{\bf AE-MSR.} We freeze the AE sub-network and train the AE-MSR network. To demonstrate the benefit of our model, we reproduce and evaluate the state-of-the-art multi-modality speech recognition model provided by~\cite{Triantafyllos-avsr2018} at different noise levels. The training begins with one-word samples, and then the length of the training sequence gradually grows. This is a cumulative method that not only improves the convergence rate on the training set, but also reduces overfitting significantly. Output size of decoder is set to 41, accounting for the 26 characters in the alphabet, the 10 digits, and tokens for [PAD], [EOS], [BOS] and [SPACE]. We also use teacher forcing method~\cite{Triantafyllos-avsr2018}, in which the ground truth of the previous decoding step serves as input to the decoder.
\begin{table}[t]
\begin{center}
\begin{tabular}{|l|c|}
\hline
{\bf Methods} & {\bf Word accuracy} \\
\hline\hline
Chung and Zisserman~\cite{chung2017lip} & 76.2\% \\
\hline
Stafylakis and Tzimiropoulos~\cite{stafylakis2017combining} & 83.0\% \\
\hline
Petridis and Stafylakis~\cite{petridis2018end} & 82.0\% \\
\hline
{\bf Ours} & 84.8\%\\
\hline
\end{tabular}
\end{center}
\caption{Word accuracy of different word-level classification networks on the LRW dataset.}
\label{table_classify}
\end{table}\\
\begin{table}[t]
\renewcommand\tabcolsep{2pt}
\begin{center}
\scalebox{0.93}{
\begin{tabular}{cccccc}
\toprule
{\bf Method}&Google~\cite{chiu2018state} & \multicolumn{2}{c}{TM-seq2seq~\cite{Triantafyllos-avsr2018}} & \multicolumn{2}{c}{EG-seq2seq}\\
\midrule
&\multicolumn{5}{c}{{\bf WER} \%}\\
\cmidrule(l){2-6}

\diagbox{{\bf SNR} dB}{\bf M}&A&A&V&A&V \\
\cmidrule(l){2-2} \cmidrule(l){3-4} \cmidrule(l){5-6}
clean&10.4&9.0&59.9&{\bf 7.2}&{\bf 57.8}\\
10&-&35.9&-&{\bf 35.5}&-\\
5&-&49.0&-&{\bf 42.6}&-\\
0&70.3&60.5&-&{\bf 58.2}&-\\
-5&-&87.9&-&{\bf 86.1}&-\\
-10&-&100.0&-&100.0\%&-\\
\bottomrule
\end{tabular}}
\end{center}
\caption{Word error rates (WER) of both single modality speech recognition and multi-modality speech recognition (MSR) on the LRS3-TED dataset. {\bf M:} modality. {\bf A:} audio modality only; {\bf V:} visual modality only.}
\label{avsr-wer}
\end{table}
\begin{table*}[t]
\renewcommand\tabcolsep{2.3pt}
\begin{center}
\scalebox{0.95}{
\begin{tabular}{c|cc|cccc|cccc}
\toprule
\multicolumn{1}{c}{\bf Modality}& \multicolumn{2}{c}{AV} &\multicolumn{4}{c}{VA}& \multicolumn{4}{c}{VAV}\\
\midrule

\multicolumn{1}{c}{\diagbox{{\bf SNR} dB}{\bf Met}}&\multicolumn{1}{c}{TM-s2s}&\multicolumn{1}{c}{EG-s2s}&\multicolumn{1}{c}{1D-TM-s2s}&\multicolumn{1}{c}{T-TM-s2s}&\multicolumn{1}{c}{1D-EG-s2s}&\multicolumn{1}{c}{T-EG-s2s}&\multicolumn{1}{c}{1D-TM-s2s}&\multicolumn{1}{c}{T-TM-s2s}&\multicolumn{1}{c}{1D-EG-s2s}&\multicolumn{1}{c}{T-EG-s2s} \\
\midrule
clean&8.0\%&6.8\%&-&-&-&-&-&-&-&-\\
10&33.4\%&32.2\%&25.9\%&24.1\%&24.2\%&{\bf 23.2}\%&24.5\%&22.0\%&21.5\%&{\bf 20.7}\%\\
5&38.1\%&36.8\%&34.1\%&31.7\%&32.7\%&{\bf 30.9}\%&30.2\%&25.6\%&26.3\%&{\bf 24.3}\%\\
0&44.3\%&41.1\%&37.0\%&33.2\%&36.6\%&{\bf 32.5}\%&31.6\%&29.6\%&28.5\%&{\bf 25.5}\%\\
-5&56.2\%&52.6\%&50.2\%&49.5\%&49.3\%&{\bf 46.0}\%&36.7\%&35.1\%&32.7\%&{\bf 31.1}\%\\
-10&60.9\%&57.9\%&52.5\%&49.8\%&50.6\%&{\bf 44.5}\%&42.3\%&42.0\%&40.2\%&{\bf 38.6}\%\\
\bottomrule
\end{tabular}}
\end{center}
\caption{Word error rates (WER) of both audio speech recognition (ASR) with single visual modality awareness and multi-modality speech recognition (MSR) with double visual modality awareness on the LRS3-TED dataset. {\bf Met:} method. {\bf TM-s2s:} TM-seq2seq; {\bf EG-s2s:} EG-seq2seq; {\bf 1D-TM-s2s:} an AE-MSR model, which consists of 1DRN-AE and TM-seq2seq; {\bf T-TM-s2s:} an AE-MSR model, which consists of TCN-AE and TM-seq2seq; {\bf 1D-EG-s2s:} an AE-MSR model, which consists of 1DRN-AE and EG-seq2seq; {\bf T-EG-s2s:} an AE-MSR model, which consists of TCN-AE and EG-seq2seq. {\bf AV:} multi-modality with single visual modality awareness; {\bf VA:} enhanced audio modality by single visual awareness for ASR; {\bf VAV:} multi-modality by double visual awareness for multi-modality speech recognition (MSR).}
\label{ae-avsr-wer}
\end{table*}

{\bf Implementation details.} The implementation of the network is based on the TensorFlow library~\cite{Abadi2016TensorFlow} and trained on a single Tesla P100 GPU with 16GB memory.  We use the ADAM optimiser to train the network with dropout and label smoothing. An initial learning rate is set to 10$^{-4}$, and decreased by a factor of 2 every time if the training error did not improve, the final learning rate is 5$\times10^{-6}$. Training of the entire network takes approximately 15 days, including the training of the audio enhancement sub-network on both of the two temporal convolution blocks and the MSR sub-network separately and the subsequent joint training. Please see our code $\footnote{\url{https://github.com/JackSyu/Discriminative-Multi-modality-Speech-Recognition}}$ for more details. 

\section{Experimental results}
\subsection{P3D-based visual front-end and EleAtt-GRU-based enc2dec}\label{5.1}
{\bf P3D-based visual front-end.} We perform lip reading experiments on both word-level and sentence-level. In section~\ref{training_strategy}, we introduce a word-level lip reading network on the LRW dataset to classify 500 word classes to train the visual front-end of C3D-P3D. Result of this word-level lip reading network is shown in Table~\ref{table_classify}, where we report word accuracy as evaluation metric and our result surpasses the state-of-the-art~\cite{stafylakis2017combining} on the LRW dataset. It demonstrates that visual front-end network of C3D-P3D is more advantageous in extracting video feature representations than the C3D-2D-ResNet one used by ~\cite{Triantafyllos-avsr2018}. 

{\bf EleAtt-GRU-based enc2dec.} Results in both of Column V and A in Table~\ref{avsr-wer} demonstrate that EleAtt-GRU-based enc2dec is more beneficial in speech recognition than the Transformer-based enc2dec. As shown in Table~\ref{avsr-wer} Column V, our multi-modality speech recognition (EG-seq2seq) network (illustrated in Figure~\ref{fig:architecture_MSR}) with only visual modality reduces word error rate (WER) by 2.1\% compared to the previous state-of-the-art (TM-seq2seq)~\cite{Triantafyllos-avsr2018} WER of 59.9\% on the LRS3 dataset without using language model in decoder. Furthermore, we also evaluate the EleAtt-GRU-based enc2dec model in ASR at different noise levels. As shown in Table~\ref{avsr-wer} Column A, EG-seq2seq exceeds the state-of-the-art (TM-seq2seq) model on ASR at all noise levels (-10 dB to 10 dB) without extra language model. Table~\ref{avsr-wer} Column A also shows that neither EG-seq2seq or TM-seq2seq works any more with only audio modality at -10 dB SNR.

Results in the columns under AV in Table~\ref{ae-avsr-wer} demonstrate the speech recognition accuracy improvement after adding the visual awareness once at the MSR stage, especially in noisy environments. Even when the audio is clean, visual modality can still play a helping role, for example the WER is reduced from 7.2\% for audio modality only to 6.8\% for multi-modality. EG-seq2seq outperforms the state-of-the-art (TM-seq2seq) model on MSR at different noise levels. It again demonstrates the superiority of EleAtt-GRU-based enc2dec in speech recognition. However, we notice that under very noisy conditions, audio modality can negatively impact the MSR because of its highly polluted input, when comparing lip reading (V in Table~\ref{avsr-wer}) with MSR (AV in Table~\ref{ae-avsr-wer}) at -10 dB SNR.  

\subsection{Audio enhancement (AE) with the first visual awareness}\label{experiment_wlc}
In order to demonstrate the enhancement effectiveness of our AE models so that it can benefit not only our speech recognition models but also other speech recognition models. Compared with MSR in the Section~\ref{5.1}, here we apply visual awareness at audio enhancement stage, instead of at MSR. We compare and analyze the results of following
networks at different noise levels:
\begin{itemize}
\item {\bf 1DRN-TM-seq2seq:} an AE-MSR network, where the audio enhancement (AE) sub-network (1DRN-AE) uses 1D ResNet as temporal convolution unit and outputs enhanced audio modality. The MSR sub-network of this network is TM-seq2seq.
\item {\bf TCN-TM-seq2seq:} an AE-MSR network, where the AE sub-network (TCN-AE) uses the temporal convolutional network (TCN) as temporal convolution unit. The MSR sub-network is TM-seq2seq.
\item {\bf 1DRN-EG-seq2seq:} an AE-MSR network, where the AE sub-network is 1DRN-AE and the MSR sub-network is EG-seq2seq.
\item {\bf TCN-EG-seq2seq:} an AE-MSR network, where the AE sub-network is TCN-AE and the MSR sub-network is EG-seq2seq.
\end{itemize}

\begin{figure}[t]
\centering
\includegraphics[width=1.0\linewidth]{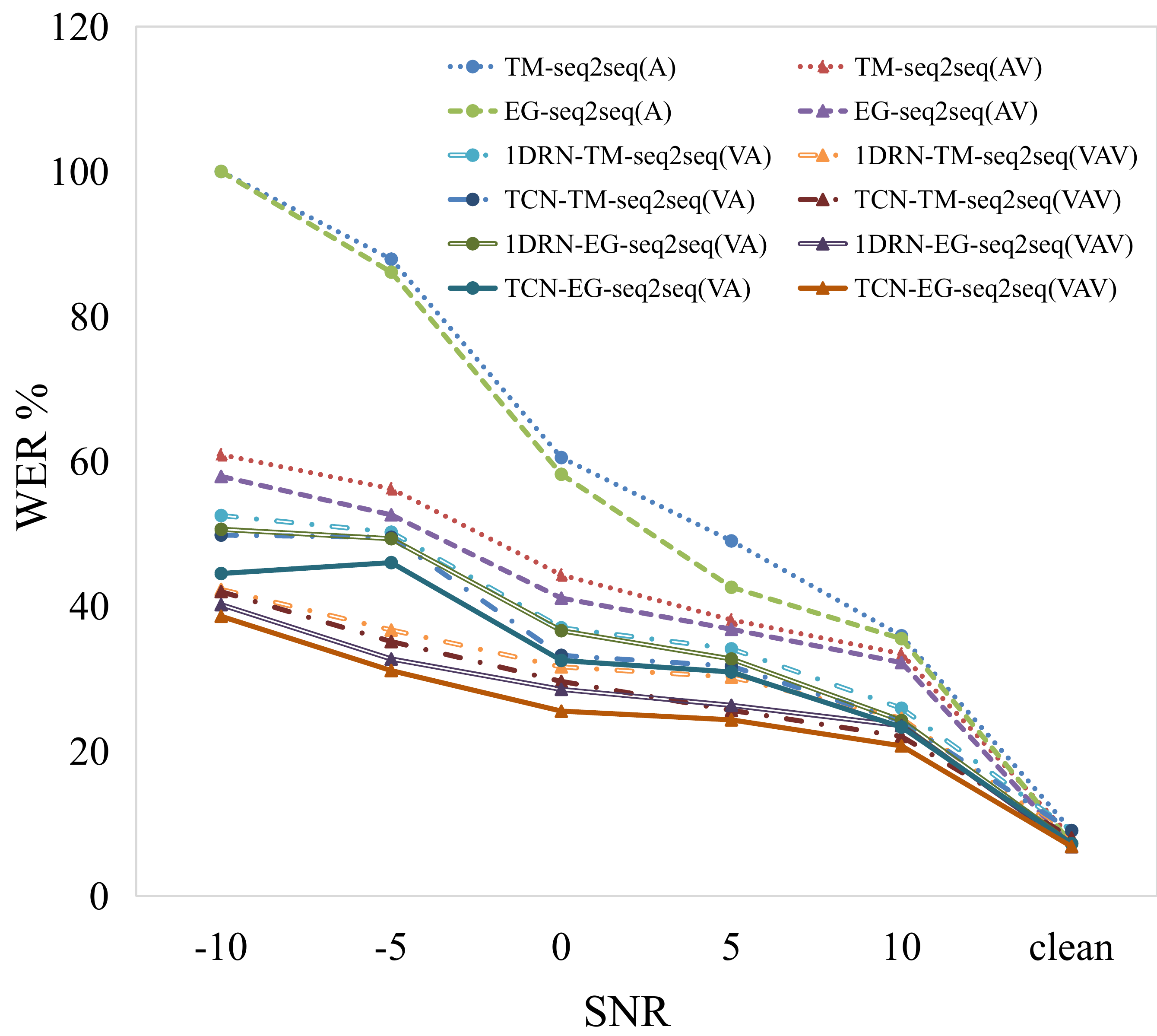}
\caption{Word error rate (WER) on different methods. Each method in this diagram equivalent to the one with same name in Table~\ref{avsr-wer}.}
\label{WER}
\end{figure}

\begin{table}[t]
\begin{center}
\begin{tabular}{ccccc}
\toprule
{\bf Audio source}& &\multicolumn{3}{c}{\bf Magnitude error \%}\\
\midrule

& SNR dB&-5&0&5\\
\cmidrule(l){2-5}
Noisy &&97.1&65.4&49.1 \\
1DRN-AE &&66.5&51.0&35.6 \\
TCN-AE &&59.5&46.3&33.1\\
\bottomrule
\end{tabular}
\end{center}
\caption{Energy errors between original noise-free audio magnitudes and enhanced magnitudes produced by different audio enhancement models.}
\label{mahgnitude_subtraction}
\vspace{-6pt}
\end{table}
In this section, all the models above use only audio modality at MSR stage. As shown in columns under VA in Table~\ref{ae-avsr-wer}, our AE networks can benefit other speech recognition models, for example at SNR of -5 dB, the WER is reduced from 87.9\% of TM-seq2seq to 50.2\% of 1DRN-TM-seq2seq and 49.5\% of TCN-TM-seq2seq.
The enhancement gain is also clearly illustrated in Figure~\ref{WER}. Moreover, by comparing the result of columns under AV and VA in Table~\ref{ae-avsr-wer}, with the same number of visual awareness, our audio enhancement approach shows more benefit in speech recognition than the multi-modality with single visual awareness in noisy environments.   

Magnitudes produced by the two AE models are shown in Figure~\ref{fig:visualization}. We also introduce an energy error function to measure the effect of audio enhancement models as follow: 
\begin{equation}
    \Delta M = \frac{\: \parallel M - {M_{o}}\parallel _{2}}{\parallel M_{o} \parallel _{2}}
\label{E1}
\end{equation}
where $M$ is the magnitudes of noisy audio or enhanced audio, $M_{o}$ is the original audio without mixing
noises, $\Delta M$ is the deviation results between $M$ and $M_{o}$. We chose 10,000 noise-free samples that are added to babble noises with SNR of -5 dB, 0 dB and 5 dB separately to compare the enhancement performance between 1DRN-AE and TCN-AE networks. We average the $\Delta M$ results among samples at each SNR-level. Results in Table~\ref{mahgnitude_subtraction} show the beneficial performance of TCN-AE. 

In Table~\ref{avsr-example}, we list some of the many examples where the single modality (video or audio alone) fails to predict the correct sentences, but these sentences are correctly deciphered by applying both modalities. It also shows that, in some noisy environment the multi-modality also fails to produce the right sentence, however the enhanced audio modality predict successfully. Experimental results of speech recognition in Table~\ref{Multi-modality speech recognition} also demonstrate that TCN-EG-seq2seq is more advantageous than 1DRN-EG-seq2seq in audio modality enhancement due to the TCN temporal convolution unit, which has a longer-term memory and larger receptive field by combining causal convolution and dilated convolution that more beneficial in temporal tasks. 

\subsection{Multi-modality speech recognition with the second visual awareness}
After audio enhancement with the first visual awareness, we implement multi-modality speech recognition with the second visual awareness. By comparing the results in columns under VA and VAV in Table~\ref{ae-avsr-wer}, MSR with double visual awareness leads to a further improvement compared to any single visual awareness method (\eg AV, VA and V). For example, the WER of 1DRN-EG-seq2seq is reduced from 36.6\% to 28.5\% when combining the visual awareness again for speech recognition after audio enhancement, and the TCN-EG-seq2seq model reduces the WER even more. It demonstrates the performance gain because of the second visual awareness in MSR. Our AE-MSR network shows significant advantage in terms of performance after combining visual awareness twice, once for audio enhancement and the other for MSR. In Table~\ref{avsr-example} we list some examples that the multi-modality model ({\bf AV}) and the AE model ({\bf VA}) fail to predict the correct sentences, but the AE-MSR model deciphers the words successfully in some noisy environments. 



\section{Conclusion}

In this paper, we introduce a two-stage speech recognition model named double visual awareness multi-modality speech recognition (AE-MSR) network, which consists of the audio enhancement (AE) sub-network and the multi-modality speech recognition (MSR) sub-network. By extensive experiments, we demonstrate the necessity and effectiveness of double visual awareness for MSR, and our method leads to a significant performance gain on MSR especially in noisy environments. Furth er, our models in this paper outperform the state-of-the-art ones on the LRS3-TED and the LRW datasets by a significant margin.
{\small
\bibliographystyle{ieee_fullname}
\bibliography{egbib}
}
\clearpage
\pagestyle{empty}
\renewcommand{\thesection}{\Alph{section}}
\setcounter{section}{0}
\section{Blocks of the Pseudo-3D (P3D) network}\label{P3D}

\begin{figure}[h]
    \centering
    \begin{subfigure}[t]{1in}
        \includegraphics[width=0.9\linewidth]{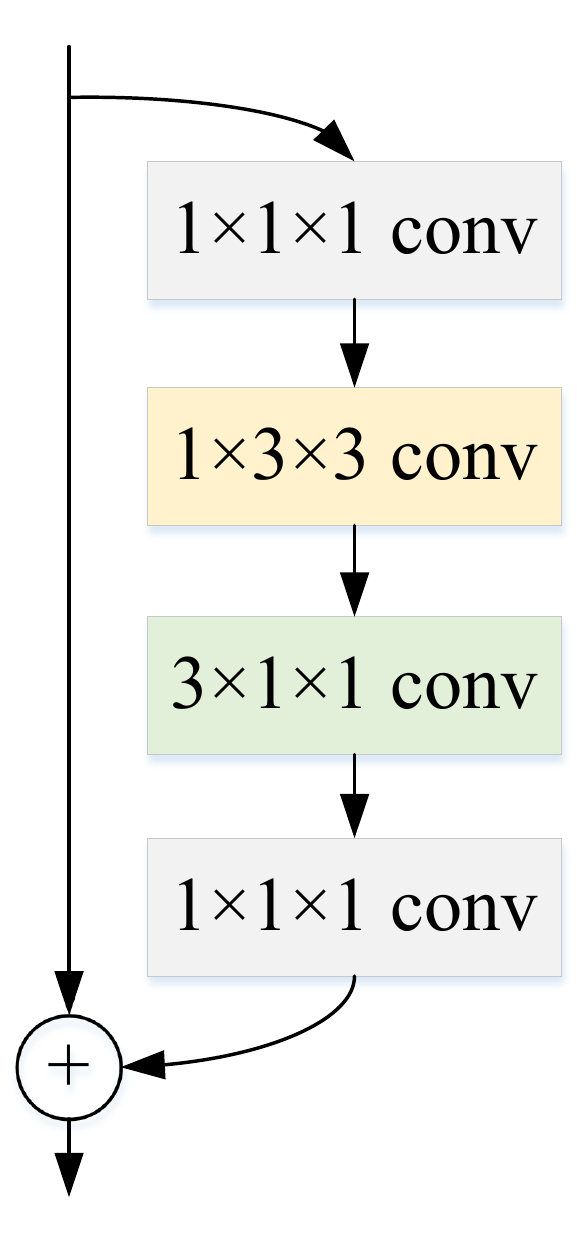}
        \caption{P3D-A}\label{fig:P3D-a}
    \end{subfigure}
    \begin{subfigure}[t]{1in}
        \includegraphics[width=1.05\linewidth]{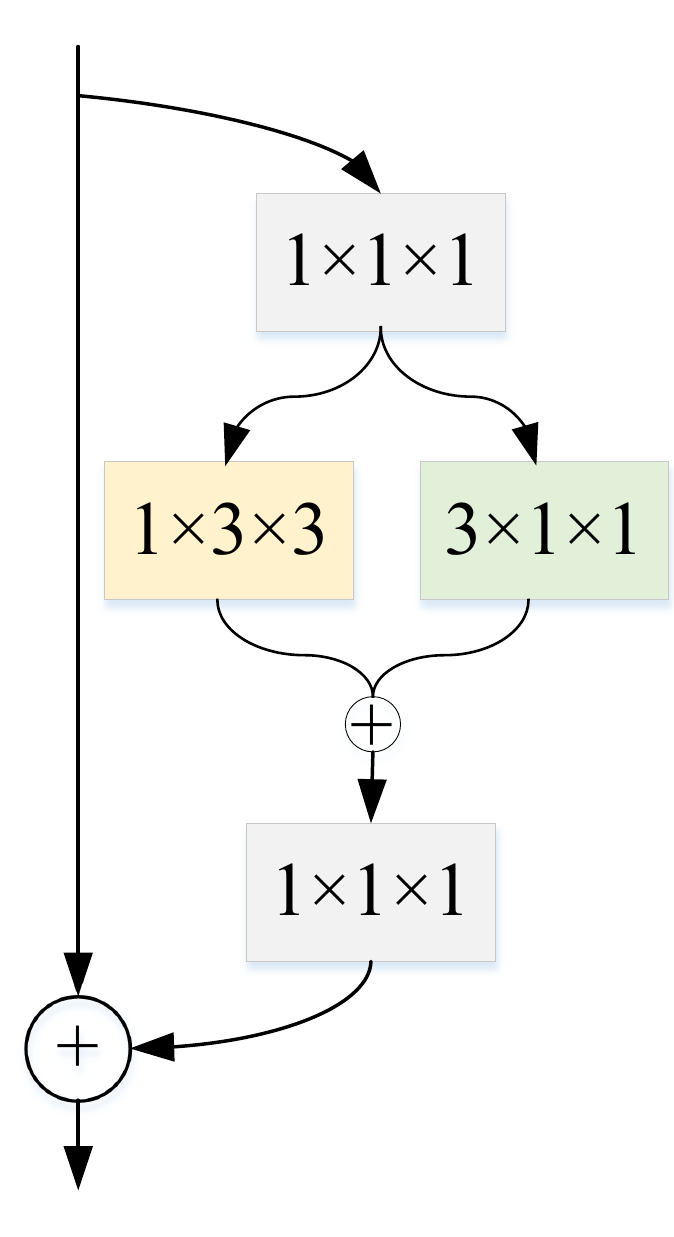}
        \caption{P3D-B}\label{fig:P3D-b}
    \end{subfigure}
    \quad
    \begin{subfigure}[t]{1in}
        \centering
        \includegraphics[width=1.0\linewidth]{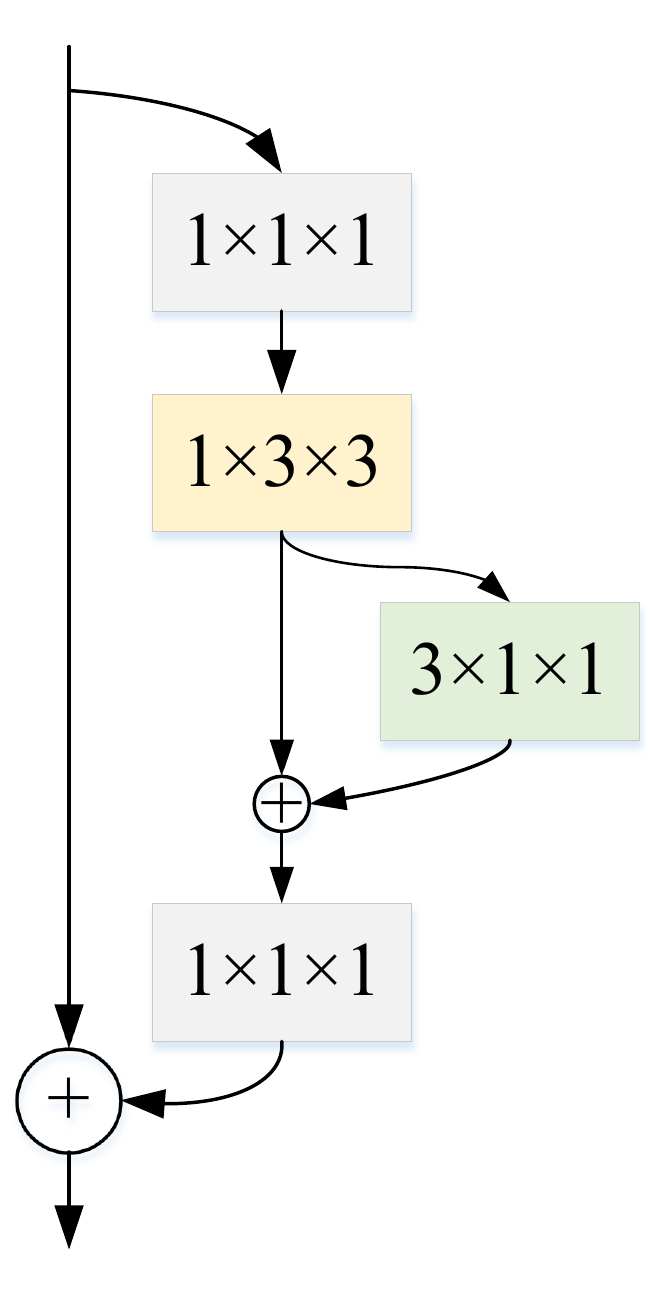}
        \caption{P3D-C}\label{fig:P3D-c}
    \end{subfigure}
         \caption{Bottleneck building blocks of the Pseudo-3D (P3D) ResNet network. P3D ResNet is produced by interleaving P3D-A, P3D-B and P3D-C in turn.}
\label{fig:P3D}
\vspace{-16pt}
\end{figure}

P3D ResNet is implemented by separating $N\times N\times N$ convolutions into $1\times3\times3$ convolution filters on spatial domain and $3\times1\times1$ convolution filters on temporal domain to extract spatial-temporal features~\cite{qiu2017learning}. The three versions of P3D blocks are shown in Figure~\ref{fig:P3D}.
\section{Details of an EleAtt-GRU block}\label{EleAtt-GRU}
\begin{figure}[h]
\includegraphics[width=1.0\linewidth]{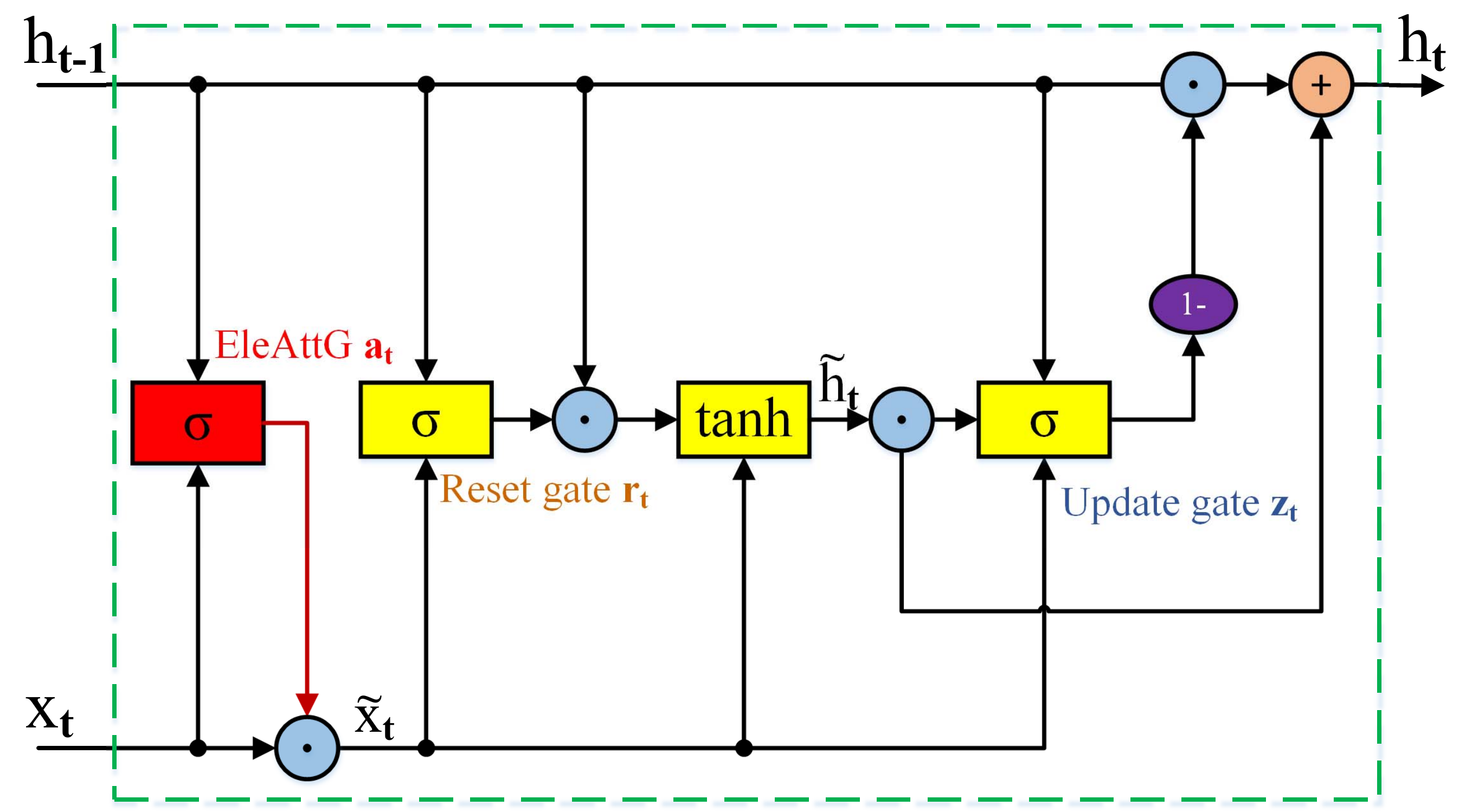}
\caption{An Element-wise-Attention Gate (EleAttG) of GRU block.}
\vspace{-15pt}
\label{fig:EleAtt-GRU}
\end{figure}

The details of the EleAtt-GRU~\cite{zhang2019eleatt} building block used by our models are outlined in Figure~\ref{fig:EleAtt-GRU}. Each GRU block has ($e.g.$, $N$) GRU neurons. {\bf Yellow} boxes -- the units of the original GRU with the output dimension of $N$. {\bf Blue circle} -- element-wise operation and the brown circle denotes vector addition operation. {\bf Red box} -- EleAttG with an output dimension of $D$, which is the same as the dimension of the input $x_{t}$.

Corresponding computations of an EleAtt-GRU are as follows:
\begin{gather}
    \tilde{x_{t}}= a_{t}\odot x_{t}\notag\\
    {r_{t}} = \sigma \left (W_{xr} \tilde{x_{t}}+W_{hr}h_{t-1}+b_{r}\right)\notag\\
    {z_{t}} = \sigma \left (W_{xz} \tilde{x_{t}}+W_{hz}h_{t-1}+b_{z}\right)\notag\\
    {{h_{t}}}' = \mathrm{tanh} \left (W_{xh} \tilde{x_{t}}+W_{hh}\left ( r_{t}\odot h_{t-1}\right )+b_{h}\right)\notag\\
    {h_{t}} = {z_{t}}\odot {h_{t-1}} + \left (1-z_{t} \right)\odot{{h_{t}}}'\notag
\end{gather}
where $\sigma$ denotes the activation function of Sigmoid. The attention response of an EleAttG is the vector $a_{t}$ with the same dimension as the input $x_{t}$ of GRU. $a_{t}$ modulates $x_{t}$ to generate $\tilde{x_{t}}$. $r_{t}$ and $z_{t}$ denote the reset gate and update gate of GRU. $h_{t}$ and $h_{t-1}$ are the output vectors of the current hidden state and the previous hidden state. $W_{\alpha \beta }$ denotes the weight matrix related with $\alpha$ and $\beta$, where $\alpha \in \left \{ x, h \right \}$ and $\beta \in \left \{ r, z, h \right \}$. $b_{\gamma}$ is the bias vector, where $\gamma \in \left \{ r, z, h \right \}$~\cite{zhang2019eleatt}.
\section{Examples of AE and AE-MSR speech recognition results.}\label{AE-MSR}
Examples of AE and AE-MSR speech recognition results are illustrated in Table~\ref{avsr-example}.
\begin{table}[h]\footnotesize
\renewcommand\tabcolsep{1.5pt}
\begin{center}
\begin{tabular}{cccc}
\toprule
& {\bf Transcription} & $\Delta {\bf M}  \%$& {\bf WER \%} \\
\midrule
{\bf GT} & \textcolor{red}{We can prevent the worst case scenario} &-&\\
{\bf V} & We can put and worst case scenario &-& 34 \\
{\bf A} & We can prevent the worst case tcario &-& 8 \\
{\bf AV} & We can prevent the worst case scenario &-& 0 \\
\midrule
&{\bf Noisy (5 dB)}&&\\
{\bf GT} & \textcolor{red}{what would that change about how we live} &-&\\
{\bf V} & wouldn't at chance whole a life &-& 53 \\
{\bf A} & that would I try all we live &36& 50 \\
{\bf AV} & that would I chance all how we live &24& 25 \\
{\bf VA (1DRN)} & what would that change about how we live &11& 0 \\
\midrule
&{\bf Noisy (0 dB)}&&\\
{\bf GT} & \textcolor{red}{human relationships are not efficient}  &-&\\
{\bf V} & you man relation share are now efficient &-& 38 \\
{\bf A} & man went left now fit &80& 73 \\
{\bf AV} & you man today are now efficient &89& 43 \\
{\bf VA (1DRN)} & human relations are now efficient &31& 14 \\
{\bf VAV (1DRN)} & human relationships are not efficient &21& 0 \\
\midrule
&{\bf Noisy (0 dB)}&&\\
{\bf GT} & \textcolor{red}{we really don't walk anymore}  &-&\\
{\bf V} & we aren't working&-& 61 \\
{\bf A} & wh ae lly son't tank  &63& 50 \\
{\bf AV} & we alley won't work more&63& 32 \\
{\bf VA (1DRN)} & we really won't work anymore &39& 11 \\
{\bf VA (TCN)} & we really don't walk anymore &22& 0 \\
\midrule
&{\bf Noisy (-5 dB)}&&\\
{\bf GT} & \textcolor{red}{at some point I'm going to get out}  &-&\\
{\bf V} & I soon planning to get it &-& 47 \\
{\bf A} & it so etolunt &96& 76 \\
{\bf AV} & at soon pant talking to get it &96& 35 \\
{\bf VAV (1DRN)} & at some point I'm taking to get out &33& 9 \\
{\bf VAV (TCN)} & at some point I'm going to get out &20& 0 \\
\bottomrule
\end{tabular}
\end{center}
\caption{Examples of recognition results by our models. {\bf GT:} Ground truth; {\bf V:} visual modality only; {\bf A:} audio modality only; {\bf AV:} multi-modality with single visual modality awareness; {\bf VA:} enhanced audio modality by single visual awareness for ASR; {\bf VAV:} multi-modality by double visual awareness for multi-modality speech recognition (MSR); {\bf 1DRN, TCN:} the temporal convolutional unit is 1D ResNet or TCN.}
\label{avsr-example}
        
\end{table}
\vspace{-6pt}
\begin{figure}[t]
\begin{center}
\includegraphics[width=1.0\linewidth]{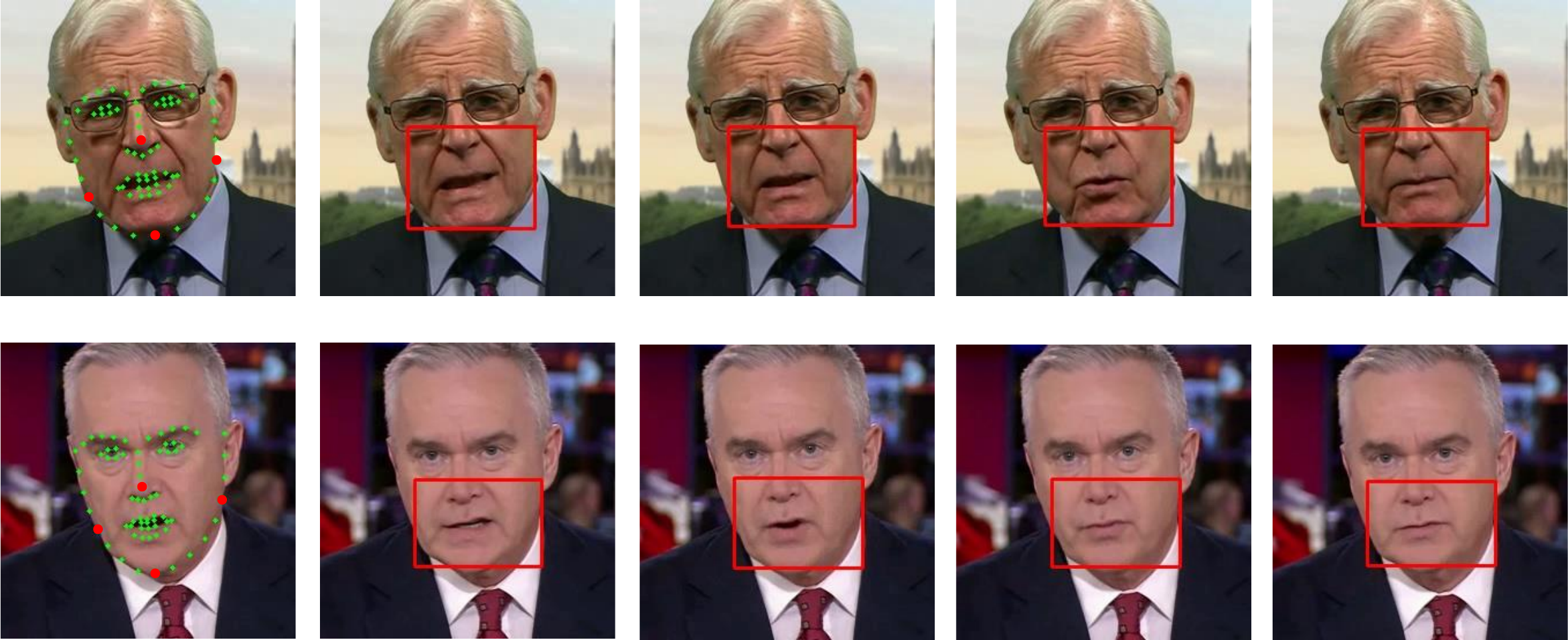}
\end{center}
\caption{Examples of mouth crop. }
\label{fig:mouth_crop}
\vspace{-2pt}
\end{figure}
\section{Enhancement examples of the 1DRN-AE and the TCN-AE models}\label{enhancement}
\begin{figure}[t]
    \centering
    \begin{subfigure}[t]{0.47\textwidth}
        \centering
        \includegraphics[width=1.0\linewidth]{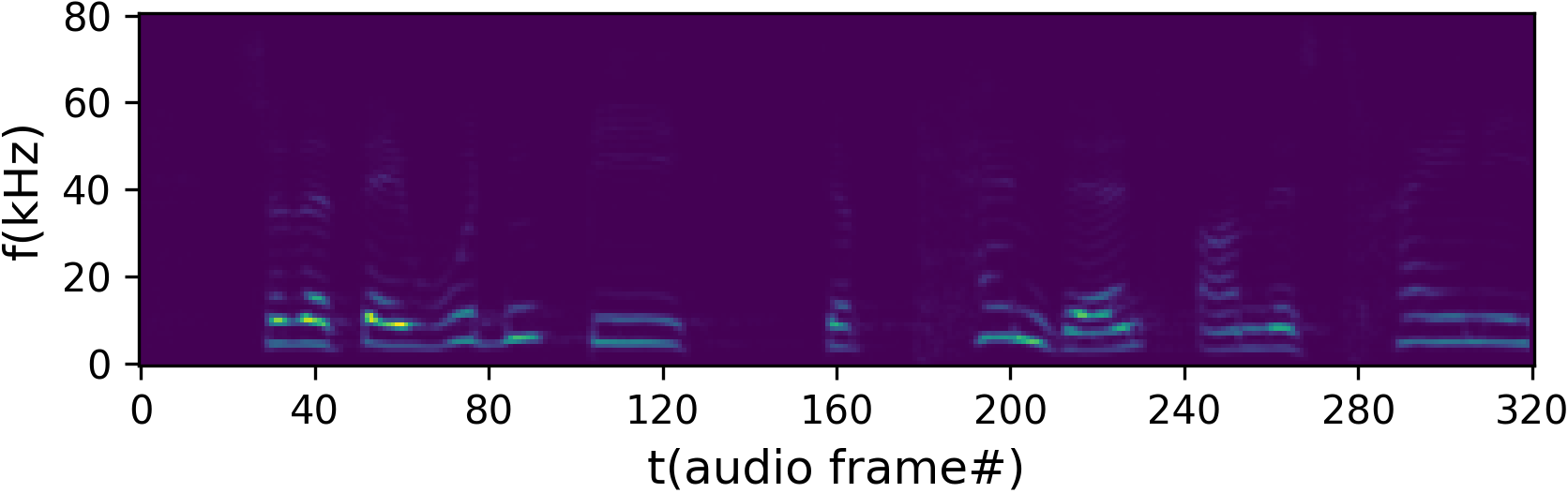}
        \caption{Clean audio magnitude}\label{fig:5a}
    \end{subfigure}
    \qquad 
    \begin{subfigure}[t]{0.47\textwidth}
        \centering
        \includegraphics[width=1.0\linewidth]{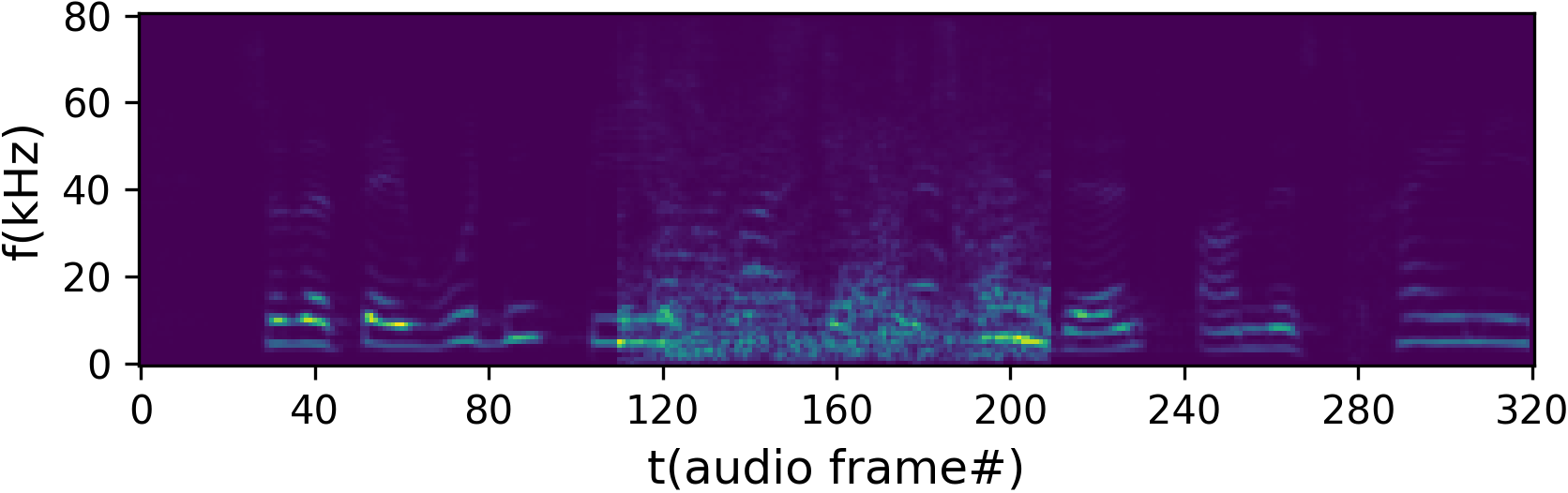}
        \caption{Noisy audio magnitude}\label{fig:5b}
    \end{subfigure}
    \quad 
    \begin{subfigure}[t]{0.47\textwidth}
        \centering
        \includegraphics[width=1.0\linewidth]{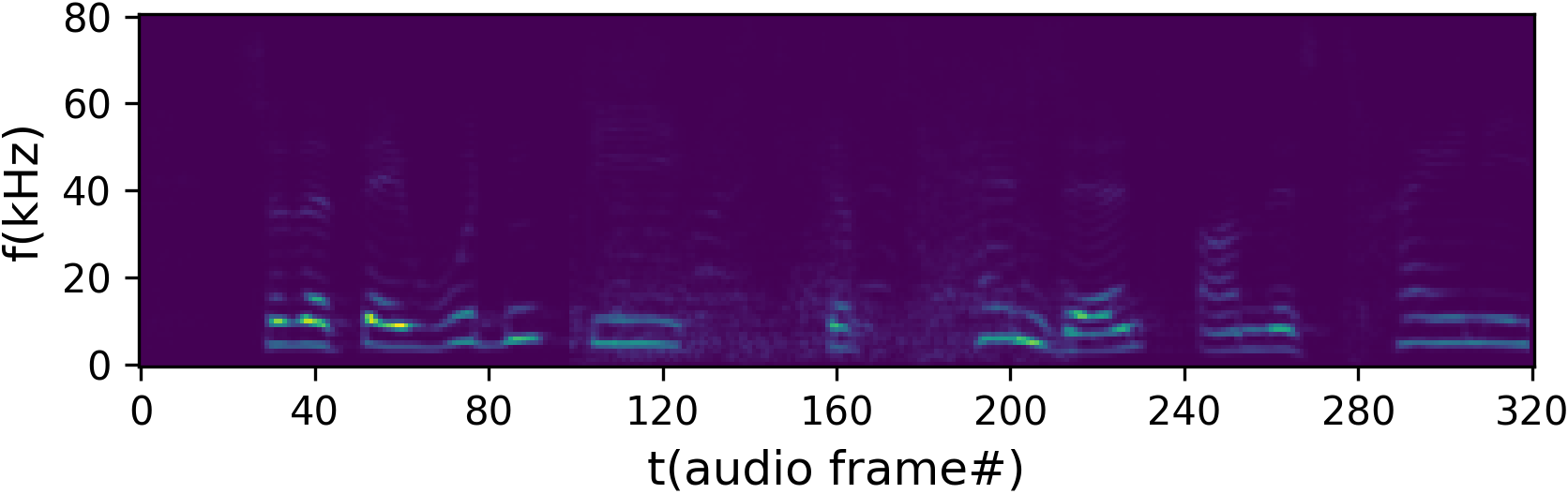}
        \caption{Enhanced audio magnitude by 1DRN-AE}\label{fig:5c}
    \end{subfigure}
    \qquad
    \begin{subfigure}[t]{0.47\textwidth}
        \centering
        \includegraphics[width=1.0\linewidth]{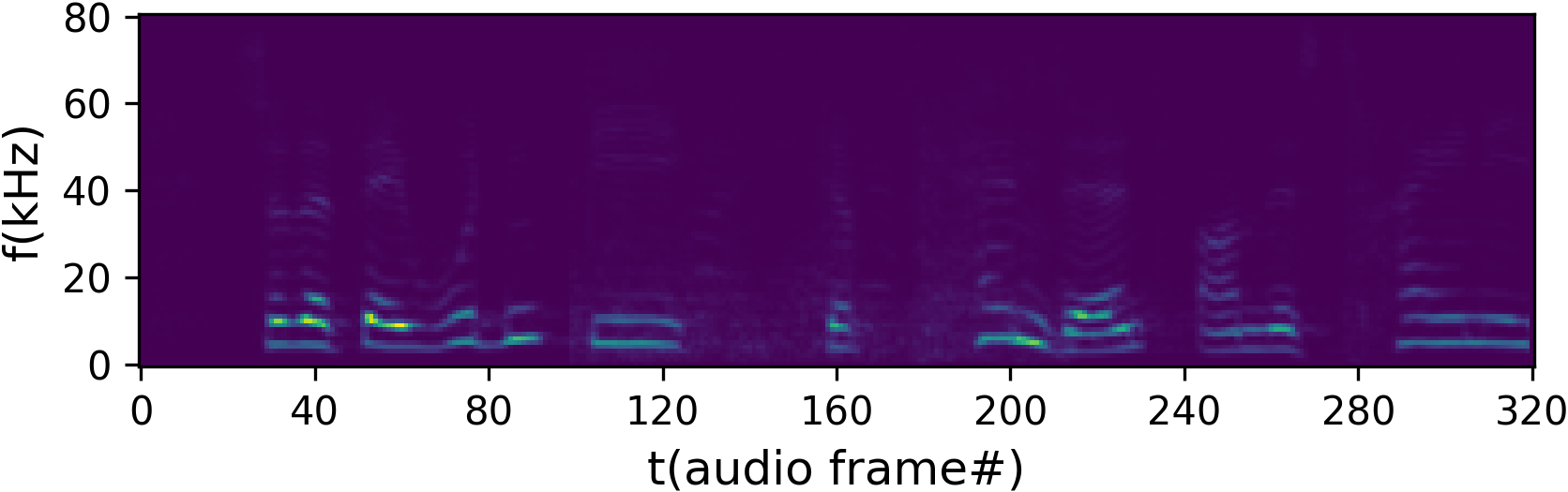}
        \caption{Enhanced audio magnitude by TCN-AE}\label{fig:5a}
    \end{subfigure}
\caption{Enhancement effects of the 1D-ResNet-based audio enhancement (1DRN-AE) model and the TCN-based audio enhancement (TCN-AE) model: {\bf a)} clean audio utterance; {\bf b)} we obtain this noisy utterance by adding babble noise to the 100 central audio frames; {\bf c)} the enhanced audio utterance by 1DRN-AE; {\bf d)} the enhanced audio utterance by TCN-AE; {\bf c)} and {\bf d)} show the effect of audio enhancement when compared to {\bf b)}.}
\label{fig:visualization}
\end{figure}

Enhancement examples of our audio enhancement sub-networks are illustrated in Figure~\ref{fig:visualization}.
\section{Examples of mouth crop}\label{mouth_crop}
We produce image frames by cropping original video frames to $112 \times 112$ pixel patches and choose mouth patch as region of interest (ROI). As shown in Figure~\ref{fig:mouth_crop}, facial landmarks are extracted by the $ Dlib $~\cite{king2009dlib} toolkit and the mouth ROI inside the red squares are achieved by 4 (red points) specified out of 68 facial landmarks (green points).
\section{Architecture details of the audio enhancement networks}\label{architecture_AE}
Architecture details of the audio enhancement sub-network are given in Table~\ref{enhancement_details}.   
\begin{table*}[t]
\begin{center}
    \begin{subtable}{2in}
        \centering
        \begin{tabular}{cccccc}
        \toprule
        {\bf Layer} & {\bf \# filters} & {\bf K} & {\bf S} & {\bf P} & {\bf Out}\\
        \midrule
        fc0 & 1536 & 1 & 1 & 1 & $T$\\
        conv1 & 1536 & 5 & 1 & 2 & $T$\\
        conv2 & 1536 & 5 & 1 & 2 & $T$\\
        conv3 & 1536 & 5 & $\frac{1}{2}$ & 2 & $2T$\\
        conv4 & 1536 & 5 & 1 & 2 & $2T$\\
        conv5 & 1536 & 5 & 1 & 2 & $2T$\\
        conv6 & 1536 & 5 & 1 & 2 & $2T$\\
        conv7 & 1536 & 5 & $\frac{1}{2}$ & 2 & $4T$\\
        conv8 & 1536 & 5 & 1 & 2 & $4T$\\
        conv9 & 1536 & 5 & 1 & 2 & $4T$\\
        fc10 & 256 & 1 & 1 & 1 & $4T$\\
        \bottomrule
        \end{tabular}
        \caption{Video Stream of 1D ResNet.}
    \end{subtable}
    \qquad\qquad
    \begin{subtable}{2in}
        \centering
        \begin{tabular}{cccccc}
        \toprule
        {\bf Layer} & {\bf \# filters} & {\bf K} & {\bf S} & {\bf P} & {\bf Out}\\
        \midrule
        fc0 & 1536 & 1 & 1 & 1 & $4T$\\
        conv1 & 1536 & 5 & 1 & 2 & $4T$\\
        conv2 & 1536 & 5 & 1 & 2 & $4T$\\
        conv3 & 1536 & 5 & 1 & 2 & $4T$\\
        conv4 & 1536 & 5 & 1 & 2 & $4T$\\
        conv5 & 1536 & 5 & 1 & 2 & $4T$\\
        fc6 & 256 & 1 & 1 & 1 & $4T$\\
    \bottomrule
    \end{tabular}
    \caption{Audio Stream of 1D ResNet.}
    \end{subtable}
    \\
  
    \begin{subtable}{2in}
        \centering
        \begin{tabular}{cccccc}
        \toprule
        {\bf Layer} & {\bf Hidden} & {\bf K} & {\bf N} & {\bf S} & {\bf Out}\\
        \midrule
        fc0 & 520 & 1 & 1 & 1 & $4T$\\
        TCN1 & 520 & 3 & 3 & 1 & $4T$\\
        fc2 & 256 & 1 & 1 & 1 & $4T$\\
        \bottomrule
        \end{tabular}
    \caption{Audio stream of TCN.}
    \end{subtable}
    \qquad\qquad    
    \begin{subtable}{2in}
        \centering
        \begin{tabular}{cccccc}
        \toprule
        {\bf Layer} & {\bf Hidden} & {\bf K} & {\bf N} & {\bf S} & {\bf Out}\\
        \midrule
        fc0 & 520 & 1 & 1 & 1 & $T$\\
        TCN1 & 520 & 3 & 3 & 1 & $T$\\
        conv2 & 520 & 3 & 1 & $\frac{1}{2}$ & $2T$\\
        TCN3 & 520 & 3 & 3 & 1 & $T$\\
        conv4 & 520 & 3 & 1 & $\frac{1}{2}$ & $4T$\\\
        fc5 & 256 & 1 & 1 & 1 & $4T$\\
        \bottomrule
        \end{tabular}
    \caption{Video Stream of TCN.}
    \end{subtable}
    \\
    \begin{subtable}{2in}
        \centering
        \begin{tabular}{ccc}
        \toprule
        {\bf Layer} & {\bf \# filters} & {\bf Out}\\
        \midrule
        EleAtt-GRU & 512 & $4T$\\
        fc1 & 600 & $4T$\\
        fc2 & 600 & $4T$\\
        fc\_mask & F & $4T$\\
        \bottomrule
        \end{tabular}
    \caption{AV Fusion.}
    \end{subtable}
    \qquad
    \begin{subtable}{2in}
        \centering
        \begin{tabular}{cccccc}
        \toprule
        {\bf Layer} & {\bf \# filters} & {\bf K} & {\bf S} & {\bf P} & {\bf Out}\\
        \midrule
        fc0 & 1536 & 1 & 1 & 1 & $4T$\\
        conv1 & 1536 & 5 & 2 & 2 & $2T$\\
        EleAtt-GRU & 128 & - & - & - & $2T$\\
        conv2 & 1536 & 5 & 2 & 2 & $T$\\
        fc6 & 512 & 1 & 1 & 1 & $T$\\
        \bottomrule
        \end{tabular}
        \caption{Enhanced audio stream.}
    \end{subtable}
    \qquad
\end{center}
\caption{Architecture details. {\bf a)} The 1D ResNet module of video stream that extracts the video features. {\bf b)} The 1D ResNet module of audio stream that extracts the noisy audio features. {\bf c)} The TCN module of video stream that extracts the video features. {\bf d)} The TCN module of audio stream that extracts the noisy audio features. {\bf e)} The EleAtt-GRU and FC layers that process multi-modality fusion and enhancing encoding. {\bf f)} The EleAtt-GRU and 1D ResNet layers that extracts the enhanced audio features.  {\bf K:} Kernel width; {\bf S:} Stride -- fractional strides denote transposed convolutions; {\bf P:} Padding; {\bf Out:} Temporal dimension of the layer’s output. {\bf Hidden:} the number of hidden units; {\bf N:} the number of TCN blocks.} 
\label{enhancement_details}
\end{table*}

\end{document}